%% file: main.tex
\pdfoutput=1

\documentclass[11pt]{article}

\usepackage[final]{acl}

\usepackage{times}
\usepackage{latexsym}

\usepackage[T1]{fontenc}

\usepackage[utf8]{inputenc}

\usepackage{microtype}

\usepackage{inconsolata}

\usepackage{graphicx}
\usepackage{subcaption}

\title{VAQUUM: Are Vague Quantifiers Grounded in Visual Data?}
\author{Hugh Mee Wong, Rick Nouwen, Albert Gatt \\
        Utrecht University \\
        Utrecht, The Netherlands \\
        \texttt{\{h.m.wong, r.w.f.nouwen, a.gatt\}@uu.nl}
        }

\usepackage{xspace}
\usepackage{booktabs}
\usepackage{bm}
\usepackage{amsmath}
\usepackage{enumitem}
\usepackage{array}
\usepackage{listings}
\usepackage{multirow}
\usepackage{multicol}
\usepackage{float}

\newcommand{\dataset}{VAQUUM\xspace}

\begin{document}
\maketitle
\begin{abstract}
Vague quantifiers such as {\em a few} and {\em many} are influenced by various contextual factors, including the number of objects present in a given context.
In this work, we evaluate the extent to which vision-and-language models (VLMs) are compatible with humans when producing or judging the appropriateness of vague quantifiers in visual contexts.
We release a novel dataset, VAQUUM, containing 20,300 human ratings on quantified statements across a total of 1089 images. Using this dataset, we compare human judgments and VLM predictions using three different evaluation methods. Our findings show that VLMs, like humans, are influenced by object counts in vague quantifier use. However, we find significant inconsistencies across models in different evaluation settings, suggesting that \emph{judging} and \emph{producing} vague quantifiers rely on two different processes.
We release our dataset and code at \url{https://github.com/hughmee/vaquum}.
\end{abstract}

\input{sections/introduction.tex}

\input{sections/related_work.tex}

\input{sections/dataset.tex}

\input{sections/exp1_logprobs.tex}

\input{sections/exp2_generation.tex}

\input{sections/exp3_mc.tex}

\input{sections/discussion.tex}

\bibliography{refs}

\appendix
\input{appendix/appendix_demographics}
\input{appendix/appendix_lmms}

\input{appendix/appendix_exp1}

\input{appendix/appendix_exp2_prompts}
\input{appendix/appendix_exp3}

\input{appendix/appendix_licenses}

\end{document}

%% file: sections/introduction.tex
\section{Introduction}
\begin{figure}[t]
    \centering
    \includegraphics[width=\linewidth]{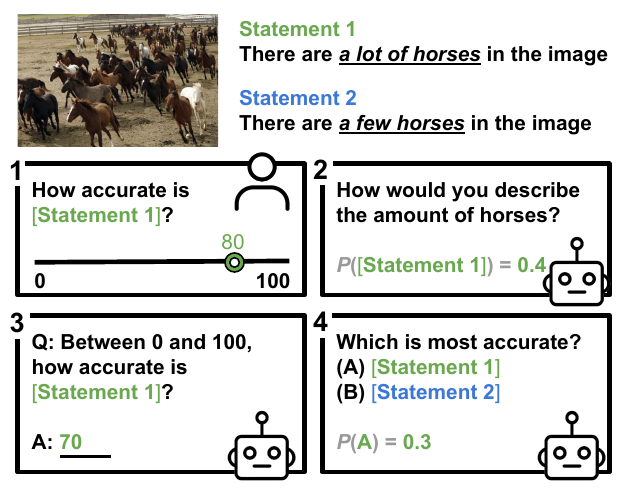}
    \caption{\textbf{Experiments in this work.} We (\textbf{1}) ask human participants to rate, using a slider, the appropriateness of statements containing vague quantifiers in relation to images. We (\textbf{2}) extract VLM generation probabilities for those same statements, (\textbf{3}) prompt the models to generate an accuracy score for them, and (\textbf{4}) evaluate probabilities assigned to these statements in a multiple-choice setup. The image above is originally from the FSC-147 dataset \citep{ranjan2021-fsc147}.}
    \label{fig:intro_example}
\end{figure}

Everyday conversations are replete with statements containing vague quantifiers, such as ``There are {\em many} horses'' (Figure~\ref{fig:intro_example}). Despite the fact that they are vague, they cause surprisingly little misunderstanding among interlocutors \citep{jucker2003-vagueness}. 
Vague quantifiers, unlike \emph{crisp} quantifiers, allow for borderline cases in which it is unclear whether the quantifier applies or not, and where we can also expect some variation in the extent to which speakers would use it. 
For example, \textit{all} does not allow for borderline cases, but it is unclear when a quantity ceases to be \textit{a few} or how many \textit{many} is.
Although vague quantifiers have long been a subject of investigation among formal semanticists \citep[see e.g.][]{nouwen2010-quantifier} and (psycho)linguists \citep[e.g.][]{moxey1993-book,deemter2010-vagueness}, they have received relatively little attention within the field of natural language processing (NLP).

In visually grounded settings, the use of vague quantifiers can be influenced by factors related to the scene itself, such as the number of entities observed \citep[e.g.][]{coventry2005-grounding}; their sizes \citep{hormann1983-calculating,coventry2010-space}; as well as information like the speaker's and hearer's personal beliefs and attitudes \citep{moxey2000-quantities,jucker2003-vagueness}. This broad range of factors, coupled with their vagueness,
raises the question of how well multimodal language models can capture human patterns in the grounded comprehension and use of such expressions. In this paper, we explore this question with vision-and-language models (VLMs) in multimodal settings involving quantified statements about images, using a novel dataset.
Studying alignment between human and VLM judgments can reveal the extent to which such models grasp the semantics of natural language.

The inclusion of a vision modality allows us to provide context in the form of both visual and textual information \citep{zhang2024-vlsurvey,ghosh2024-frontier}.
We examine to what extent visual cues influence state-of-the-art VLMs' understanding and production of expressions containing vague quantifiers, and how this compares to human linguistic intuitions (Figure~\ref{fig:intro_example}). 
By zooming in on such visual and contextual variables, 
our work follows the spirit of recent research exploring the grounding capabilities of VLMs
\cite[e.g.][]{zellers2019-visualcommonsense,thrush2022-winoground,zhang2022-commonsense,parcalabescu2022-valse,chen_bla_2023,kamath_hard_2024,wang_mementos_2024}.
The contributions of this paper are as follows.

\begin{itemize}
    \item We release VAQUUM (\textbf{Va}gue \textbf{Qu}antifiers with H\textbf{um}an Judgments), a new dataset pairing images of different types of objects with their counts, as well as human judgments of different quantified statements corresponding to the image.
    While designed for our research objectives, the dataset’s wider utility is discussed in Section~\ref{sec:discussion}.
    
    \item We analyze the features of the visual context that influence both human and model judgments on the appropriateness of different vague quantifiers, including counts, the segmentation area occupied by the target objects, and aspects of world knowledge such as their normative size.

    \item We show that VLMs do, to some extent, follow human patterns in judging the appropriateness of vague quantifiers. However, the behavior of models and their degree of alignment with human judgments depend on the evaluation paradigm used (Figure~\ref{fig:intro_example}): approaches that rely on extracting probabilities for quantified statement or that allow the model to choose the most appropriate statement, yield better alignment than methods which prompt the models to numerically rate the statements for their appropriateness given an image.
\end{itemize}

%% file: sections/related_work.tex
\section{Related Work}
The use and judgment of vague quantifiers have been studied extensively in formal semantics and psycholinguistics. 
Recent years have also seen a growing but relatively limited interest in studying (V)LM behavior with linguistic quantifiers. 

\paragraph{Vague quantifiers in formal semantics}
Work on vague quantifiers in formal semantics sits at the crossroads of generalized-quantifier theory and degree semantics. The foundational work of \citet{barwise_generalized_1981} drew attention to properties of natural language quantifiers such as monotonicity and entailment; this serves as the basis for subsequent work extending generalized quantifier theory to vague quantifiers \cite[e.g.][]{partee_many,fernando_kamp,cohen_relative_2001}. A parallel line of work has focused on the treatment of vagueness in natural language, e.g.\ in gradable adjectives. Here, an important insight is that vague predicates are frequently interpreted with respect to a standard that is set by some salient frame of reference \cite[e.g.][{\em i.a.}]{klein_80,graff_interest,Kennedy_vagueness_grammar,deemter2010-vagueness}. 
\citet{solt_vagueness_2011} shows that comparison classes (frames of reference) also play an important role in the interpretation of vague quantifiers such as \textit{many} and \textit{few}.

\paragraph{Vague quantifiers in psycholinguistics}
It has been suggested that humans make use of an \emph{approximate number system} \citep{feigenson2004-number, dehaene2011-numbersense, coventry2005-grounding}, where vague terms might not refer to exact numbers but rather approximations thereof.
Related to this, \emph{subitizing} is the ability to instantly and accurately recognize a small number of items without going through the process of counting.
Humans are generally able to do this for object counts up to 4 \citep{kaufman1949, mandler1982-subitizing}.
Above this subitizability threshold, they tend to use quantifiers \citep{barr2013-quantified, berger2023-descriptions}.
However, it has also been shown that quantifier comprehension and use go beyond (approximations of) the cardinality of the targeted object. 
Factors include object size \citep{hormann1983-calculating,newstead2000-context}; 
the number and proportions of \emph{other} objects in the scene \citep{coventry2005-grounding,coventry2010-space,pezzelle2018-mental};
set size (e.g.\ the answer to a question such as: ``\emph{Several} marbles from a set of 12 marbles would be \rule{0.3cm}{0.15mm} marbles''; \citealp{newstead1987-set});
the functionality of objects in the scene \citep{newstead2000-context}; and object grouping and spacing \citep{coventry2005-grounding}.

In conversations and texts, the choice of quantifier influences the expected rhetorical impact of a statement, and vice versa. \citet{moxey1993-prior} show that the choice of quantifier can reveal a speaker's \emph{prior expectations} regarding the frequency of the object in the scene. Moreover, several works have outlined the different perspectives that \textit{a few} and \textit{few} convey: while ``\emph{a few} people were at the party'' focuses on those who were present, ``\emph{few} people were at the party'' puts the emphasis on those who did not attend \citep{moxey2000-quantities,paterson2009-discourse}.

\paragraph{(Vague) quantifiers in NLP}
Most work on evaluating VLMs on quantifiers has focused on \emph{crisp} quantifiers (e.g.\ \textit{none}, \textit{all} and \textit{more than half}) rather than vague ones. \citet{sorodoc2016-quantify} show that neural networks can be trained to learn the quantifiers \textit{no}, \textit{some} and \textit{all} without the need for an explicit counting system. \citet{sorodoc2018-neural} extend this to a visual question-answering (VQA) task with natural images. They include vague expressions with \textit{few} and \textit{some}, but define these terms using specific proportions (e.g.\ {\em few} applies for predications involving less than 17\% of objects in the domain).
A similar definition is adapted by \citet{pezzelle2017-fuzzy}, who show that models require different mechanisms for learning cardinals and quantifiers.
Note that once the range of a quantifier is defined, it can no longer be considered \emph{vague} as borderline cases are excluded.

Moving beyond the gold label approach, \citet{testoni2019-sound} demonstrate that models using both audio and visual input to select appropriate quantifiers can achieve results that align with human distributions reported by \citet{pezzelle2018-mental}. 
\citet{enyan2024-quantifiers} compare human and large language model (LLM) responses on questions such as ``There are 500 balls. 234 of them are yellow. Are many balls yellow?''
They find that responses generated by LLMs align more closely with human judgments on crisp quantifiers than on vague ones. 
\citet{belem2024-uncertainty} find that LLMs can map uncertainty expressions such as \textit{probably} and \textit{unlikely} to probabilistic (numerical) responses in a human-like fashion.
More akin to our experiments, \citet{testoni2024-quantifying} evaluate three VLMs on their abilities to assign appropriate quantifiers to visual scenes, prompting models to select one out of nine quantifiers in response to questions such as ``How many animals are there in the image?'', with 
synthetic images generated by \citet{pezzelle2018-mental}. 
Our approach diverges from theirs on several points. 
First, we use natural images rather than artificial ones, offering a more realistic setting for evaluating VLMs.
Additionally, we use a wider range of methods to provide a more comprehensive assessment of model behavior.

%% file: sections/dataset.tex
\begin{figure*}[t]
    \centering
    \includegraphics[width=\linewidth]{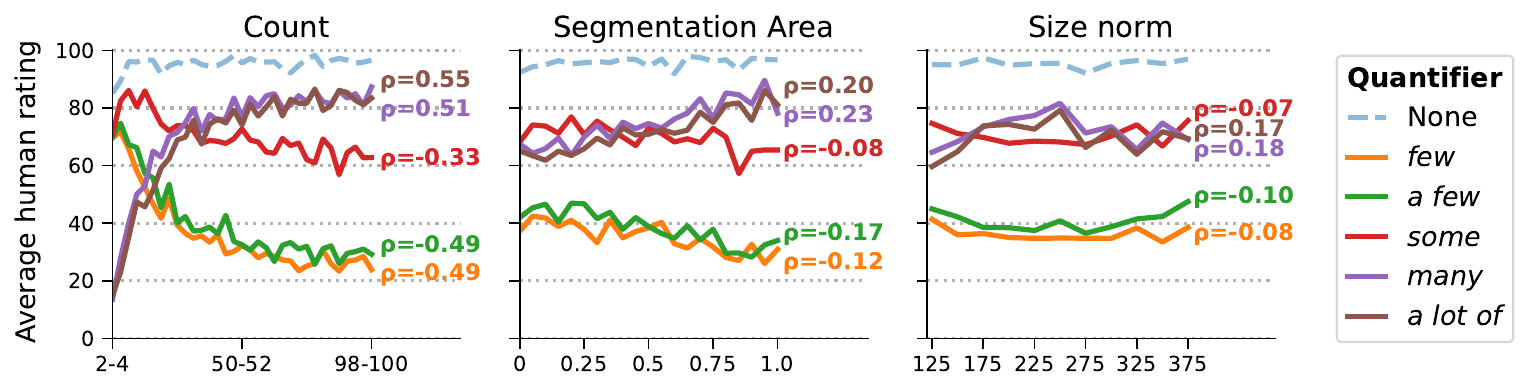}
    \caption{\textbf{Average human ratings with increasing counts, segmentation area and size norms.} For each variable and each quantifier, we also report Spearman's $\rho$, which are all statistically significant ($p <$ 0.05).}\label{fig:human_results}
\end{figure*}

\section{The \dataset Dataset}\label{sec:dataset}
We construct and release the VAQUUM dataset: \textbf{Va}gue \textbf{Qu}antifiers with H\textbf{um}an Judgments. 

\paragraph{Images}
We utilize annotated datasets used for object counting in computer vision.
FSC-147 \citep{ranjan2021-fsc147} contains 6146 images across 147 object types,
with annotated object counts ranging from 7 to 3731.
\citet{hobley2022-fsc133} refine and deduplicate this dataset to release FSC-133 (containing 133 object types).
We sample images from FSC-133 and exclude a total of 22 object categories for several reasons, 
such as their uncountable nature (e.g.\ \textit{fresh cut}), obscurity (e.g.\ \textit{carrom board pieces}) or simply because the images do not depict the object from the label. 
We also remap 37 categories to either their plural form, where necessary, or their basic-level category \cite[e.g.\ mapping \textit{crows} to \textit{birds}; cf.][]{Rosch1976}. 
Since the lowest count in FSC-133 is 7, we complement this dataset with samples from 
the test set of TallyQA \citep{acharya2019-tallyqa}, 
which includes images and annotated counts sourced from 
Visual Genome \citep{krishna2017-visualgenome} and VQA2 \citep{antol2015-vqa,goyal2017-vqa2}. 
Here, we use images classified as ``simple'' in TallyQA, 
which have counts between 1 and 15. 
From this set, we exclude images for which the labelled object is not in the set of remapped FSC-133 labels. 
We discard all counts below 2 (from TallyQA) and above 100 (from FSC-133). 
We include three types of object features in our dataset.

\paragraph{1. Count bin}
To address the imbalance in object counts within the merged dataset,
we group the 99 distinct counts (ranging from 2 to 100) into bins of three (counts from 2 to 4, 5 to 7, etc).
From each bin, we randomly sample 33 images, yielding
1089 images, evenly distributed across 33 count bins, covering counts from 2 to 100. 

\paragraph{2. Segmentation area}
We estimate the segmentation area of the object(s) in each image, i.e. the ratio of pixels in the objects' bounding region over the total image area. For each image, we prompt CLIPSeg \citep{luddecke2022-clipseg}, with the name of the object type (e.g \textit{birds}). The output logits are then passed through a sigmoid function, and the resulting values are thresholded. The resulting binary mask is used to compute the segmentation area, which essentially corresponds to ``object size'' in previous work.

\paragraph{3. Size norm}
We investigate the impact of real-world object size using the object-specific norms in the THINGSplus database \citep{stoinski2024-thingsplus}, an extension of THINGS \citep{hebart2019-things}. Such norms are collected from human judges, and they reflect ``average'' or ``typical'' values for specific properties. The \emph{size} 
norm tells us something about an object's perceived real-life size, on an arbitrary scale. 
Objects that are not explicitly present in this dataset are either mapped to the closest (base) category or discarded in our size norm analyses. 

\subsection{Human Judgments}\label{sec:dataset_humans}
We recruited 203 participants, all native and primary speakers of English, through Prolific (52.2\% female; 45.8\% male; 1.5\% undisclosed). Participant ages ranged from 25 to 84, with the majority aged 25-34 (31.5\%) and 35-44 (25.6\%). 

\subsubsection{Procedure}
We presented each participant with 100 questions in a random order. 
Each of these questions consists of an image and a statement of the form ``There are \verb|[QUANT]| \verb|[OBJECT]| in the image.'' 
Here, \verb|OBJECT| is the plural form of the object depicted and
\verb|QUANT| $\in$ \{\textit{few}, \textit{a few}, \textit{some}, \textit{many}, \textit{a lot of}\} (e.g.\ ``There are \textit{a lot of apples} in the image.''). For each image, we also include the unquantified statement (omitting \verb|QUANT|).
Participants were asked to rate, using a slider, how accurate the statement is for the image (see Figure~\ref{fig:intro_example}). The slider ranges from ``Completely inaccurate'' to ``Completely accurate''. 
No participant saw the same image twice. 

\subsubsection{Analysis}\label{sec:human_analysis}
We analyze the effects of count, segmentation area and size norms on the collected appropriateness ratings of the vague quantifiers. 
We summarize the results in Figure~\ref{fig:human_results}.

We observe from Figure~\ref{fig:human_results} that an increase in count leads to an increase in the average ratings assigned to statements containing \textit{many} and \textit{a lot of}, whose trajectories are nearly identical. 
Conversely, for the complementary pair \textit{few} and \textit{a few}, we find that average ratings \emph{decrease} as object count increases. As expected, judgments for unquantified control statements are independent of count, with the exception of a slightly lower rating for the lowest counts. We also observe that \textit{few}/\textit{a few} and \textit{many}/\textit{{a lot of}} exhibit opposing trends in relation to count, again as expected.
These observations are broadly in line with findings by e.g.\ \citet{coventry2010-space}.
Average ratings for \textit{some} also decrease as count increases, though less steeply than for (\textit{a}) \textit{few}.
While the signs of Spearman's coefficient are the same across all predictors, the strength of the correlation for segmentation area and size norm is noticeably lower. 
Furthermore, \textit{few}/\textit{a few} and \textit{many}/\textit{{a lot of}} do not exhibit opposing trends as a function of area or size norm, as they do with count.

\begin{table}
    \centering
    \setlength{\tabcolsep}{5pt}
    \small
    \begin{tabular}{l|ccccc|l}
        \toprule
        & \textbf{\textit{few}} & \textbf{\textit{a few}} & \textbf{\textit{some}} & \textbf{\textit{many}} & \textbf{\textit{a lot of}} & \textbf{ME}\\\midrule
        \textbf{C} & -0.37 & -0.38 & -0.20 & 0.38 & 0.42 & 0.03\\
        \textbf{SG} & -0.07 & -0.10 & -0.05 & 0.08 & 0.06 & 0.04 \\
        \textbf{SN} & -0.13 & -0.11 & -0.07 & 0.14 & 0.17 & 0.01$^*$ \\\midrule
        \textbf{ME} & -1.71 & -1.60 & -0.73 & -0.60 & -0.69
            \\\bottomrule
    \end{tabular}
    \caption{\textbf{Estimates of the linear mixed effects model fit to data in VAQUUM.} \textbf{C}=Count, \textbf{SG}=Segmentation, \textbf{SN}=Size norm, \textbf{ME}=Main effect.  All numbers are statistically significant ($p <$ 0.05), except the one marked (*). For main effects, the quantifier is releveled to the unquantified case, with intercept estimated at $\beta=$ 0.89.}
    \label{tab:lmm_humans}
\end{table}

To gain further insights into the relations between participants' ratings and object count and size, we fit a linear mixed effects model (LMM) to our data, predicting human judgments from the fixed effects of
quantifiers, count, segmentation area and size norm
and using participants and object category as random effects. 
We include interaction terms between pairs of predictors to investigate their joint influence on judgments.
For full details of the LMM, we refer to Appendix~\ref{appendix:lmms}. 

We report LMM estimates of the main effects and two-way interaction effects in Table~\ref{tab:lmm_humans}. 
All main effects except those for size norm are statistically significant.
For the two-way interactions, \textit{few}, \textit{a few} and \textit{some} consistently show negative estimates across all predictors, while \textit{many} and \textit{a lot of} are consistently positive. As expected given the trends in Figure~\ref{fig:human_results}, object count exhibits the strongest impact on each quantifier. 
Estimates for segmentation area and size norm display similar trends, but with weaker effects.
The LMM explains $50.3\%$ of the total variance in our participant data ($R^2c=$ 0.503, $R^2m=$ 0.459).
The random effects present moderate variability at the participant level, with a variance of 0.042
suggesting that individual differences among participants explain some of the variance in judgments.
In contrast, the object random effect accounts for minimal variance (0.002),
indicating that differences between objects have little influence on the judgments given by participants in our experiments. 

%% file: sections/exp1_logprobs.tex
\begin{figure*}[ht]
    \centering
    \includegraphics[width=0.92\linewidth]{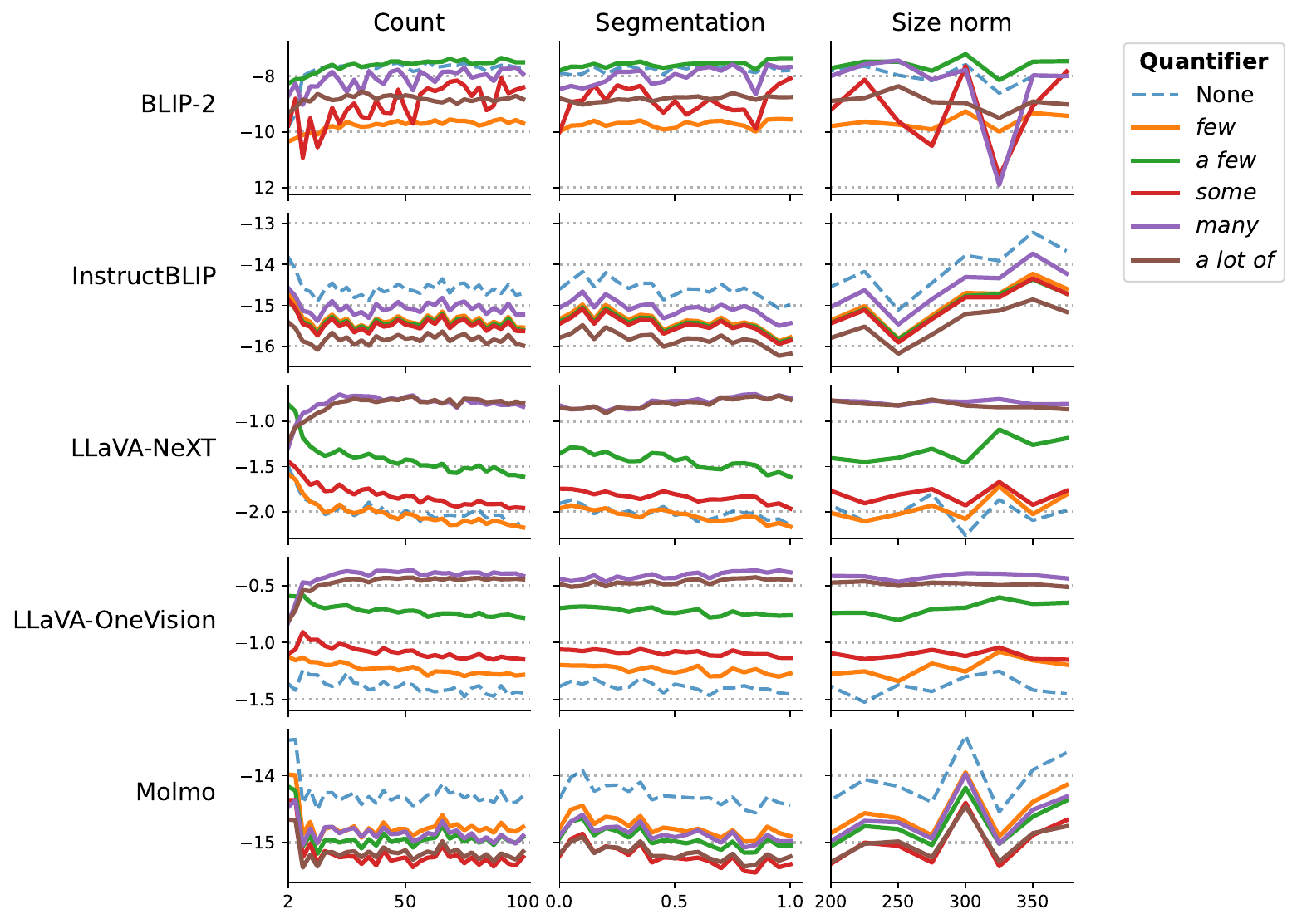}
    \caption{\textbf{Log probabilities as functions of count, segmentation area and size norm.} The patterns reported for LLaVA-NeXT and LLaVA-OneVision are most similar to human ratings. We find that InstructBLIP and Molmo do not distinguish between the quantifiers at all, whereas BLIP-2 moderately correlates with humans for \textit{many} and \textit{a lot of}.}
    \label{fig:exp1_logprobs}
\end{figure*}

\section{Experiment 1: Production Probabilities}\label{sec:logprobs}
Our first series of experiments studies the predicted \emph{production probabilities} of quantified statements by SOTA VLMs. We prompt the models with 
``How would you describe the amount of \verb|[OBJECT]| in the image?''
We extract log probabilities, conditioned on this prompt and the image, 
for the quantified statements in \dataset, as well as the unquantified version.
All extracted scores are normalized by token length. We consider the following models.

\begin{description}
    \item[BLIP-2]\citep{li2023-blip2}\textbf{.} We use the checkpoint powered by OPT-6.7B \citep{zhang2022-opt} connected to a EVA-CLIP ViT-g \citep{radford2021-clip, fang2023-eva} image encoder via a lightweight Query transformer. 
    
    \item[InstructBLIP]\citep{dai2023-instructblip}\textbf{.} We use the checkpoint with a Vicuna-13B \citep{zheng2023-vicuna} language backbone, instruction-tuned on BLIP-2.

    \item[LLaVA-NeXT]\citep{liu2024-llavanext}\textbf{.} 
    We use the 7B checkpoint with a Mistral \citep{jiang2023-mistral} language backbone.
    It integrates a CLIP-ViT through an MLP vision-language connector.

    \item[LLaVA-OneVision]\citep{li2024-llavaov}\textbf{.} We utilize the 7B checkpoint, which integrates a SigLIP \citep{zhai2023-siglip} vision encoder with a Qwen2 \citep{yang2024-qwen2} language decoder. 
     
    \item[Molmo]\citep{deitke2024-molmo}\textbf{.} We use the 7B-D checkpoint, which connects a ViT image encoder to Qwen2.7B via a connector MLP.
\end{description}

Figure \ref{fig:exp1_logprobs} displays predicted log probabilities as a function of count, segmentation area and size norm and Table~\ref{tab:exp1_correlations} reports correlations between model predictions and human judgments.

\paragraph{Alignment with humans}
Of the VLMs tested, the two LLaVA models exhibit the highest
correlation with the human data in \dataset. 
For these models, we observe in Figure~\ref{fig:exp1_logprobs} patterns similar to those of \dataset in Figure~\ref{fig:human_results}. 
Probabilities for \textit{many} and \textit{a lot of} increase as a function of count, while \textit{few} and \textit{a few} show a downward trend. Given that the question in the prompt focused explicitly on the {\em amount} of objects, the unquantified sentence is expected to be generally dispreferred. The trends in Figure~\ref{fig:exp1_logprobs} suggest
that the LLaVA models can indeed draw this distinction between quantified and unquantified statements, as the unquantified expression displays the lowest-ranking log probabilities across count, segmentation and size norm. 
However, other models do not reveal that same ability. 
This is most pronounced for InstructBLIP and Molmo, which generally tend to favor the unquantified statement as a response to the question. 
These models also show the same pattern across all quantifiers, further confirming their inability to differentiate among them. 
While the behavior of BLIP-2 is seemingly random, Figure~\ref{fig:exp1_logprobs} shows an upward trend for all quantifiers as a function of count.

\begin{table}
    \centering
    \small
    \begin{tabular}{l|ccccc}
    \toprule
    \textbf{Model} & \textbf{\textit{few}} & \textbf{\textit{a few}} & \textbf{\textit{some}} & \textbf{\textit{many}} & \textbf{\textit{a lot of}} \\ \midrule
        BLIP-2          & \textbf{-0.18} & \textbf{-0.19} & \textbf{-0.06} & \textbf{0.14} & \textbf{0.13} \\
        InstBLIP    & 0.06 & 0.04 & -0.03 & -0.01 & -0.04 \\
        LLaVA-N      & \textbf{0.34} & \textbf{0.39} & \textbf{0.21} & \textbf{0.43} & \textbf{0.52} \\ 
        LLaVA-O & \textbf{0.30} & \textbf{0.40} & \textbf{0.22} & \textbf{0.52} & \textbf{0.54} \\
        Molmo           & \textbf{0.16} & \textbf{0.20} & \textbf{0.07} & \textbf{-0.17} & \textbf{-0.21} \\
    \bottomrule
    \end{tabular}
    \caption{\textbf{Pearson's correlation between human ratings and model log probabilities.} Numbers in boldface are statistically significant ($p <$ 0.05). We provide Spearman's correlation scores in Appendix~\ref{appendix:exp1}.}
 \label{tab:exp1_correlations}
 \end{table}
 
\begin{table}
    \centering
    \small
    \setlength{\tabcolsep}{5pt}
    \begin{tabular}{l|ccccc|c}
    \toprule 
        & \textbf{\textit{few}} & \textbf{\textit{a few}} & \textbf{\textit{some}} & \textbf{\textit{many}} & \textbf{\textit{a lot of}} & \textbf{ME} \\\midrule
        \textbf{C}  & 0.00 & -0.01 & -0.02 & \textbf{0.22} & \textbf{0.22} & \textbf{-0.09} \\
        \textbf{SG} & -0.02 & -0.01 & 0.01 & \textbf{0.07} & \textbf{0.05} & \textbf{-0.05} \\
        \textbf{SN} & \textbf{0.04} & \textbf{0.05} & -0.03 & \textbf{0.12} & \textbf{0.09} & \textbf{-0.05} \\\midrule
        \textbf{ME} & \textbf{0.39} & \textbf{1.68} & \textbf{0.77} & \textbf{2.46} & \textbf{2.32}\\
        \bottomrule
    \end{tabular}
    \caption{\textbf{Estimates of the LMM for log probabilities of LLaVA-OneVision}. 
    \textbf{C}=Count, \textbf{SG}=Segmentation, \textbf{SN}=Size norm, \textbf{ME}=Main effect.  Boldface indicates statistical significance ($p <$ 0.05). For the main effects, the quantifier is releveled to the unquantified case and the estimate of the intercept is $\beta = \textbf{-1.25}$.}
    \label{tab:lmm_llavaov}
\end{table}

\paragraph{Linear mixed model}
In Table~\ref{tab:lmm_llavaov}, we display the estimates of a linear mixed effects model fit to log probabilities of LLaVA-OneVision (see Appendix~\ref{appendix:lmms} for details and Appendix~\ref{appendix:exp1} for the remaining models). 
Following our approach in \S\ref{sec:dataset}, we predict model probabilities from the fixed effects of quantifiers, count, segmentation area and size norm while including object category as a random effect. 
The latter shows a variance of 0.056, indicating that object category accounts for a moderate amount of variance among predicted log probability scores. 
Moreover, we see in Table~\ref{tab:lmm_llavaov} that \textit{many} and \textit{a lot of} show statistically significant interactions with all predictors, with the strongest effects observed with count, just as was the case for the human judgments. The estimates for the other quantifiers, however, are very different from what we found for humans. 
Overall, the LMM explains 91.2\% of the total variance in our data ($R^2m =$ 0.861, $R^2c =$ 0.912).

\paragraph{Prompts should target \emph{amounts}} 
For most models, we find that simply changing the question from 
``How would you describe \emph{the amount of} \verb|[OBJECT]| in the image?'' 
to ``How would you describe the image?'' 
yields different patterns in the results (see Appendix~\ref{appendix:exp1}). 
Most notably, we find that the observed similarity between trends in human judgments and model predictions disappears once the prompt does not focus on amounts.

\paragraph{Interim conclusion}
In \S\ref{sec:dataset_humans}, most estimates of the LMM fit to participant data were statistically significant. 
Moreover, object count made the biggest difference across all quantifiers.
For LLaVA-OneVision, the model displaying the highest Pearson's correlation with human data in Table~\ref{tab:exp1_correlations}, a similar result can be found in  Table~\ref{tab:lmm_llavaov} for \textit{many} and \textit{a lot of}: effects of interaction with object count are most pronounced, after which size norms have a slightly higher impact than segmentation area. However, these effects are absent for the other quantifiers.
BLIP-2, InstructBLIP and Molmo do not show meaningful interactions between their predicted log probabilities and the three contextual variables.

%% file: sections/exp2_generation.tex
\section{Experiment 2: Generating Judgments}\label{sec:model_judgments}
We now evaluate the instruction-tuned VLMs using an approach that is more akin to the way \dataset was constructed in \S\ref{sec:dataset}. 
That is, we prompt the models to explicitly rate the acceptability of quantified statements.
We experimented with 10 different prompts that are variations on the question shown to human participants in \S\ref{sec:dataset_humans}. 
Drawing inspiration from prompts used by \citet{belem2024-uncertainty}, we center our analyses in the remainder of this section around the following prompt: ``On a scale of 0 (completely inaccurate) to 100 (completely accurate), how accurate is the following statement for the image? Please respond with one of the following options: 0, 5, 10, 15, 20, 25, 30, 35, 40, 45, 50, 55, 60, 65, 70, 75, 80, 85, 90, 95, 100. \verb|[Statement]|'', where \verb|Statement| is an expression from the \dataset dataset. 
We refer to Appendix~\ref{appendix:exp2} for the complete list of prompts tested.

\paragraph{For VLMs, appropriateness is not gradable}
We find that in this evaluation setup, BLIP-2 and InstructBLIP generally fail to generate numerical responses to the prompts we tested, despite some prompts explicitly encouraging them to only respond with a number. 
The two LLaVA models and Molmo consistently provide numerical responses to most of the prompt templates tested. 
However, while we construct the prompts in such a way that VLMs are encouraged to provide a response that falls \emph{between} a certain range, the vast majority of model responses tend towards the extremes (i.e.\ on the lower or upper bound of the specified range; see Appendix~\ref{appendix:exp2} for a distribution of responses). 

\begin{figure}
    \centering
    \includegraphics[width=\linewidth]{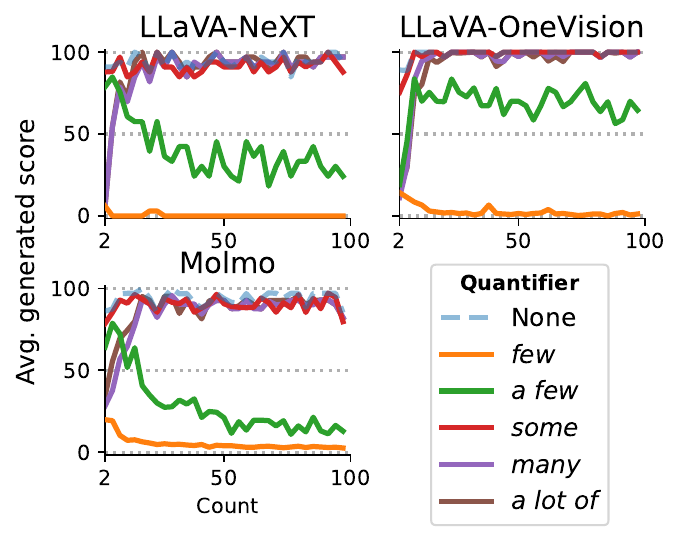}
    \caption{\textbf{Scores generated by VLMs in Experiment~2.} Note that we do not display results for BLIP-2 and InstructBLIP, as those models generally failed to provide numerical responses to the prompt.}
    \label{fig:exp2_score_generation}
\end{figure}

\paragraph{\textit{Some} is generally appropriate}
When numeric answers to prompts tend towards the extremes of a scale, it can be informative to aggregate generated scores, which is virtually the same as calculating the \emph{relative frequency} of a VLM dis/agreeing with the statements. 
We report this in Figure~\ref{fig:exp2_score_generation} for object count and make the following observations. 
First, statements containing the quantifier \textit{few} are rarely deemed appropriate.
For the models in the LLaVA family, arguably the most interesting deviation from Figure~\ref{fig:exp1_logprobs} is that in this setting, \textit{some} is considered an accurate quantifier, regardless of object count. 
Indeed, we observe that the trajectory of \textit{some} in Figure~\ref{fig:exp2_score_generation} corresponds to that of the unquantified condition.
We hypothesize that in the case of \emph{judging} the appropriateness of \textit{some}, this vague quantifier could be interpreted as an \emph{existential} quantifier. 
That is, ``There are some apples in the image'' can be regarded as a confirmation of the existence of apples in the image. 

\paragraph{Interim conclusion}
Experiment~1 showed that object count influences model predictions for \textit{many} and \textit{a lot of}.
Similar patterns emerge in Figure~\ref{fig:exp2_score_generation}, where average scores for these quantifiers increase with count.
Discrepancies between results from Experiments~1 and 2 show that in a setting where models are explicitly required to judge statements (Exp~2), the outcomes are unrelated to the models' log probabilities for the same statements (Exp~1). In Experiment~1, probabilities are extracted using an autoregressive method compatible with the pretraining objective of the LLM backbone. In contrast, Experiment~2 relies on model abilities acquired during post-training, which further modifies model parameters. The discrepancies we observe align with independent observations that post-training can negatively impact model calibration \cite{kalai_calibrated_2024,zhu_calibration_2023}.

%% file: sections/exp3_mc.tex
\begin{figure*}[ht]
    \centering
    \includegraphics[width=\linewidth]{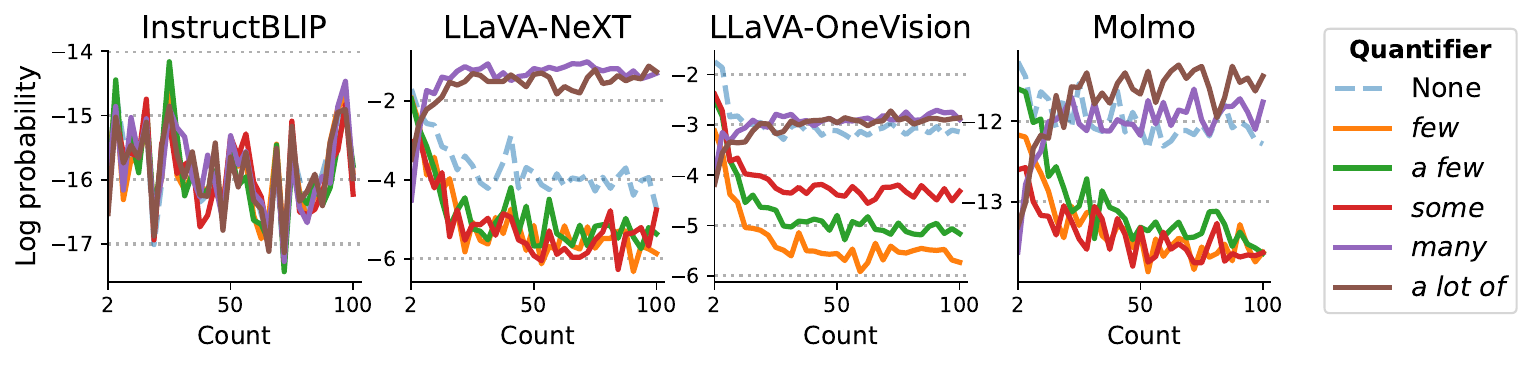}
    \caption{\textbf{Log probabilities extracted for multiple-choice labels in Experiment~3.} We do not display results for BLIP-2 because that model is not instruction-tuned.}\label{fig:exp3_mc}
\end{figure*}

\section{Experiment 3: Multiple-Choice QA}\label{sec:exp3}
Finally, we evaluate VLM judgments in a multiple-choice question-answering (MCQA) setup using a standard MCQA template of the form 
``Question: Which statement is most accurate for the image? Select the answer from the options below. \verb|[OPTIONS]| Answer: ('', 
with \verb|OPTIONS| being the set of all statements for an image in \dataset, labeled \verb|(A)| to \verb|(F)|. 
For each image, the order of the expressions is shuffled to mitigate the effects of positional biases \cite{zong_fool_2024}.
To compare the different quantifiers and ensure that the VLMs do not produce irrelevant output, we extract the log probabilities of the labels rather than allowing VLMs to generate a response.
Note that, differently from \S\ref{sec:logprobs} and \S\ref{sec:model_judgments}, the VLMs are now presented with {\em all} statements before being prompted for a response. 

In Figure~\ref{fig:exp3_mc}, we report the predicted log probabilities of \emph{instruction-tuned} VLMs as a function of count.
Table~\ref{tab:exp3_correlations} shows the correlation of these scores with both the human judgments and the log probabilities from Experiment~1.
It is clear that in this setup, too, InstructBLIP fails to differentiate between the various quantified statements. 
However, while Molmo behaved similarly in Experiment~1, it distinguishes between quantifiers in the current setting.
For Molmo and the two LLaVA models, count influences predictions for \textit{many/a lot of} and for \textit{few/a few} in the expected direction. 
This is most pronounced in the lower count ranges. 
Patterns for \textit{some} once again differ from those found in our earlier experiments. While probabilities for {\em some} generally fell between those of \textit{few} and \textit{a few} in Experiment~1, and {\em some} was generally judged appropriate in Experiment~2, we now observe that it follows the same trend as \textit{few} and \textit{a few}, while being slightly preferred over these two by LLaVA-OneVision.

\paragraph{Interim conclusion}
The two LLaVA models and Molmo show moderate correlation with human scores in \dataset.
They also correlate with their log probabilities from Experiment~1.
These models are also the most self-consistent.
While Molmo is not self-consistent, in the multiple-choice setup it correlates better with human ratings.

\begin{table}
\centering
\small
\setlength{\tabcolsep}{5pt}

\begin{tabular}{ll|ccccc}
\toprule
&& \textbf{\textit{few}} & \textbf{\textit{a few}} & \textbf{\textit{some}} & \textbf{\textit{many}} & \textbf{\textit{a lot of}}\\\midrule

\multirow{2}{*}{\rotatebox[origin=c]{90}{\textbf{INB}}}
    & $r(\textup{VAQ})$ & 0.00 & 0.00 & 0.01 & -0.01 & 0.04 \\
    & $r(\textup{EXP1})$ & \textbf{-0.13} & \textbf{-0.14} & \textbf{-0.12} & \textbf{-0.13} & \textbf{-0.15}\\
    \midrule

\multirow{2}{*}{\rotatebox[origin=c]{90}{\textbf{LLN}}} 
    & $r(\textup{VAQ})$ & \textbf{0.32} & \textbf{0.27} & \textbf{0.14} & \textbf{0.42} & \textbf{0.33}\\
    & $r(\textup{EXP1})$ & \textbf{0.36} & \textbf{0.35} & \textbf{0.26} & \textbf{0.44} & \textbf{0.35}\\
    \midrule
\multirow{2}{*}{\rotatebox[origin=c]{90}{\textbf{LLO}}} 
    & $r(\textup{VAQ})$ & \textbf{0.45} & \textbf{0.45} & \textbf{0.19} & \textbf{0.35} & \textbf{0.43}\\
    & $r(\textup{EXP1})$ & \textbf{0.33} & \textbf{0.42} & \textbf{0.24} & \textbf{0.35} & \textbf{0.42} \\
    \midrule
\multirow{2}{*}{\rotatebox[origin=c]{90}{\textbf{MOL}}} 
    & $r(\textup{VAQ})$ & \textbf{0.26} & \textbf{0.31} & \textbf{0.15} & \textbf{0.28} & \textbf{0.35} \\
    & $r(\textup{EXP1})$ & \textbf{0.25} & \textbf{0.28} & \textbf{0.25} & \textbf{-0.07} & \textbf{-0.12}
    \vspace{1.25pt}\\
\bottomrule

\end{tabular}
\caption{\textbf{Pearson's $\bm{r}$ of log probabilities in Experiment~3 with human data (VAQ) and log probabilities from Experiment~1 (EXP1)}. Models shown are InstructBLIP (\textbf{INB}), LLaVA-NeXT (\textbf{LLN}), LLaVA-OneVision (\textbf{LLO}) and Molmo (\textbf{MOL}). Boldfaced numbers are statistically significant. We display Spearman's correlation coefficients in Appendix~\ref{appendix:exp3}.}
\label{tab:exp3_correlations}
\end{table}

%% file: sections/discussion.tex
\section{Discussion}\label{sec:discussion}

\paragraph{Alignment with humans}
In this paper, we explore how vision-and-language models produce and evaluate simple expressions containing vague quantifiers. 
We constructed the \dataset dataset and used this to investigate whether object count, segmentation area and size norm affect VLMs to the same extent as they do humans. 
We showed that in particular for object count, the patterns found in some VLMs show striking similarities with the human data in \dataset. This result appears to contradict the observation that VLMs perform poorly on counting tasks \citep{parcalabescu2021-counting, parcalabescu2022-valse}. However, our findings with vague quantifiers could be accounted for in terms of an \emph{approximate number system}, which cognitive scientists have posited to account for the human ability to rapidly estimate quantities \citep{feigenson2004-number, condry2008-number, dehaene2011-numbersense,odic2018-ans,piantadosi_rational_2016}. 
In the context of vague quantifiers, it has been argued that there exists a mapping between exact and approximate number systems \citep{coventry2005-grounding, coventry2010-space}. The extent to which VLMs rely on something akin to an ANS is a topic for future work.

To gather data that better aligns with Experiment~1, one possible approach is to leverage existing human-annotated datasets from image captioning studies.
However, commonly used datasets such as MSCOCO \citep{lin2014-coco} typically feature low count ranges, leading to a scarcity of relevant images as counts increase.
Moreover, while some captions do include the quantifiers central to our study, such instances are rare: most images have at most one caption containing a relevant quantifier, and annotators often avoid using them altogether.
Nevertheless, we see potential in incorporating naturalistic datasets that include human-generated descriptions.
Expanding our work to include such resources (potentially through re-annotation with a focus on quantifiers) could offer additional answers to the question of human-model alignment on vague quantifiers.

\paragraph{Self-consistency}
Our experiments relied on paradigms incorporating \emph{production} (Experiment~1) and \emph{judgment} (Experiments~2 and 3). We find that VLMs are \emph{not} self-consistent across these evaluation paradigms. 
That is, when a VLM is set to \emph{judge} the use of a quantifier---a meta-linguistic task---its judgment is not necessarily rooted in the log probabilities that govern the model's generation of the quantifier. Questions of calibration and consistency such as these go beyond the domain of quantifiers and are an active area of ongoing research \cite[e.g.][]{krause_confidently_2023, giulianelli_what_2023, zhu_calibration_2023}.

\paragraph{Perspectivism}
In this paper, we follow the standard practice of aggregating human judgments through taking the average and focusing on the general trends. This, however, might overlook meaningful variability that emphasizes the complexity of human judgments on vague expressions.
While developed with our specific research questions in mind, the \dataset dataset captures not only quantifier judgments in visual contexts, but also the variation and disagreement among humans.
Modeling disagreement among annotators is increasingly recognized as an important focus for NLP, as it reveals the shortcomings of assuming a single ground truth (such as the opinion of the majority). This view is one of the cornerstones of
\emph{perspectivist} approaches to NLP \citep{frenda2024-perspectivist, cabitza2023-perspectivist, abercrombie2024-nlperspectives}.
We see our dataset as a useful potential resource for this community, as well as any community that studies linguistic phenomena where variation is bound to play a role.
Vagueness is a case in point, as it inherently gives rise to disagreements---not as noise, but as a result of the context-sensitive nature of language.

\paragraph{Outlook}
Psycholinguistics has shown that vague quantifiers do not depend exclusively on the count and size of target objects.
This is further confirmed by the residual variance (49.7\%) in \dataset that cannot be explained by the linear mixed effects model (LMM) on human judgments.
While the LMM analysis yields a better fit for VLM log probabilities, we find that there, too, the LMM cannot explain all the variance (leaving a residual variance of 8.8\% for LLaVA-OneVision). 
Future work could focus on other contextual factors, such as the number of \emph{other} objects present, the object density in the image, as well as the role of scene semantics and other objects in the image background. 
In combination with visual grounding capabilities, it is worthwhile to investigate the role of commonsense and world knowledge in vague quantifier usage: 
while seeing 20 people at a conference will most likely not be reason for one to exclaim that there are \emph{many}, the same amount of toddlers at such an event might be.

\section*{Limitations}


\paragraph{Model selection}
Our experiments focus on a selection of vision-and-language models.
While this selection has allowed us to compare models from the same model family (BLIP-2 and InstructBLIP; LLaVA-NeXT and LLaVA-OneVision), as well as models that share similar language model backbones (LLaVA-OneVision and Molmo), 
conclusions drawn in this study can be better generalized with experiments on a wider range of vision-and-language models. We hope that the \dataset dataset provides the impetus for further model comparisons.

\paragraph{Segmentation area and size norm}
Given that of the three contextual variables, the role of object count has been most prominent in literature on vague quantifiers, we focused on selecting images that balance a range of counts that we deemed representative. 
Estimating the segmentation area and extracting the size norms for these images may subsequently have yielded distributions that do not represent the full range of values that these variables can take on.
It is therefore possible that the distributions for segmentation area and size norm were too sparse to say something more meaningful about their roles in \dataset and model results. 
Thus, while we at times find statistically significant relationships between judgments and segmentation area or size norm, future work could focus on investigating the \emph{practical} significance.
Additionally, we recognize that using CLIPSeg to estimate the segmentation area can introduce inaccuracies.

\section*{Ethical Considerations}
The data collection for \dataset underwent an ethics check in our institution. The data collected via crowdsourcing does not contain any information that can be traced back to individuals. No materials were used to our knowledge that could harm or otherwise adversely affect individuals.

\section*{Acknowledgments}
We thank the three anonymous reviewers for their helpful comments.
We thank the members of the Natural Language Processing group at Utrecht University for their valuable suggestions on earlier versions of this paper.
We are particularly grateful to the members of the Vision-and-Language lab  and to Sandro Pezzelle for the insightful discussions.
This work is funded by the Dutch Research Council (NWO) through the AiNed Fellowship Grant NGF.1607.22.002, \textit{Dealing with Meaning Variation in NLP}.

%% file: appendix/appendix_demographics.tex
\section{Data from Human Participants}

\subsection{Instructions and Consent}
Below we include the information given to the participants in our human experiment.

\textit{Thank you for taking part in this experiment. This survey should take approximately 20 minutes to complete. You will be presented with 100 questions. Each question consists of an image and a corresponding statement. Your task is to rate, using a slider, how accurate you find the statement in relation to the image.}

\textit{Please be assured that all responses will be kept strictly confidential and anonymous. The data that we collect will be processed in such a way that they cannot be linked to you in any way. Participation in this survey is entirely voluntary. If at any point you wish to exit the survey without finishing the survey, you can close this form and we will delete your responses. You do not have to specify your reason.}

\textit{Should you wish to withdraw consent after you have participated, please send an email to \textup{\textbf{AUTHORS}} at \textup{\textbf{EMAIL}}. Note that if you withdraw consent after completing the survey, we are not required to undo the processing of your data that has taken place up until that time.
}

\textit{If you wish to participate in the study, please check the following box. If you do not wish to do so, you can close this tab.}

\subsection{Demographics}
In \S\ref{sec:dataset_humans}, we mentioned that we recruited 203 participants through Prolific.
As reported in the Ethical Considerations, we did not collect data that allows anyone to trace the responses back to an individual.
All participants were native and primary speakers of English.
We have the following additional information about the distribution of demographics.

\begin{description}
    \item[Age] 25-34 years (31.5\%), 35-44 (25.6\%), 18-24  (17.2\%), 45-54 (15.3\%), 55-64 (6.9\%), 65-74 (2.5\%) and 75-84 (0.5\%). 0.5\% of the participants prefer not to disclose their age.

    \item[Gender] female (52.5\%), male (45.8\%), other (0.5\%). 1.5\% of the participants prefer not to say.
\end{description}

\subsection{Participant reward}
Participants were found through Prolific and were paid \pounds\ 2.50 for 20 minutes (\pounds\ 7.50 per hour).

%% file: appendix/appendix_lmms.tex
\section{Linear Mixed Effects Models}\label{appendix:lmms}
Below we provide the details for the linear mixed effects models that we fit to our data.
All LMMs are fit using the \verb|lme4| package in R. 

\subsection{Human Data (\dataset)}
In \S\ref{sec:dataset}, we are interested in predicting human judgments from the main effects of quantifiers, object count, segmentation area and size norms, as well as the interaction between these predictors.
We include the participants and object categories as random effects. Put concretely, 

\begin{verbatim}
  judgment ~ quantifier * count 
           * segmentation * size_norm 
            + (1|participant) + (1|object)
\end{verbatim}

We scale judgments, count, segmentation area and size norm to make sure they all have a mean of 0 and a standard deviation of 1. For example,
\begin{verbatim}
  count <- scale(count,
                center=TRUE,
                scale=TRUE)
\end{verbatim}
\noindent
This way, we ensure that we can meaningfully interpret the relation between one unit of change in one variable with a change in another.
Additionally, we make the variables for quantifier and object category a \verb|factor| and relevel the quantifier to use the unquantified (\verb|base|) condition as the reference category.
\begin{verbatim}
  quantifier <- relevel(quantifier,
                        ref="base")
\end{verbatim}

\subsection{Model Data (Experiment 1)}
For the models, we follow the same steps taken as those for fitting an LMM to human data, but now we no longer have to account for different participants.
That is, 

\begin{verbatim}
  log_prob ~ quantifier * count 
            * segmentation * size_norm 
            + (1|object)
\end{verbatim}

%% file: appendix/appendix_exp1.tex
\begin{figure}[t]
    \centering
    \includegraphics[width=\linewidth]{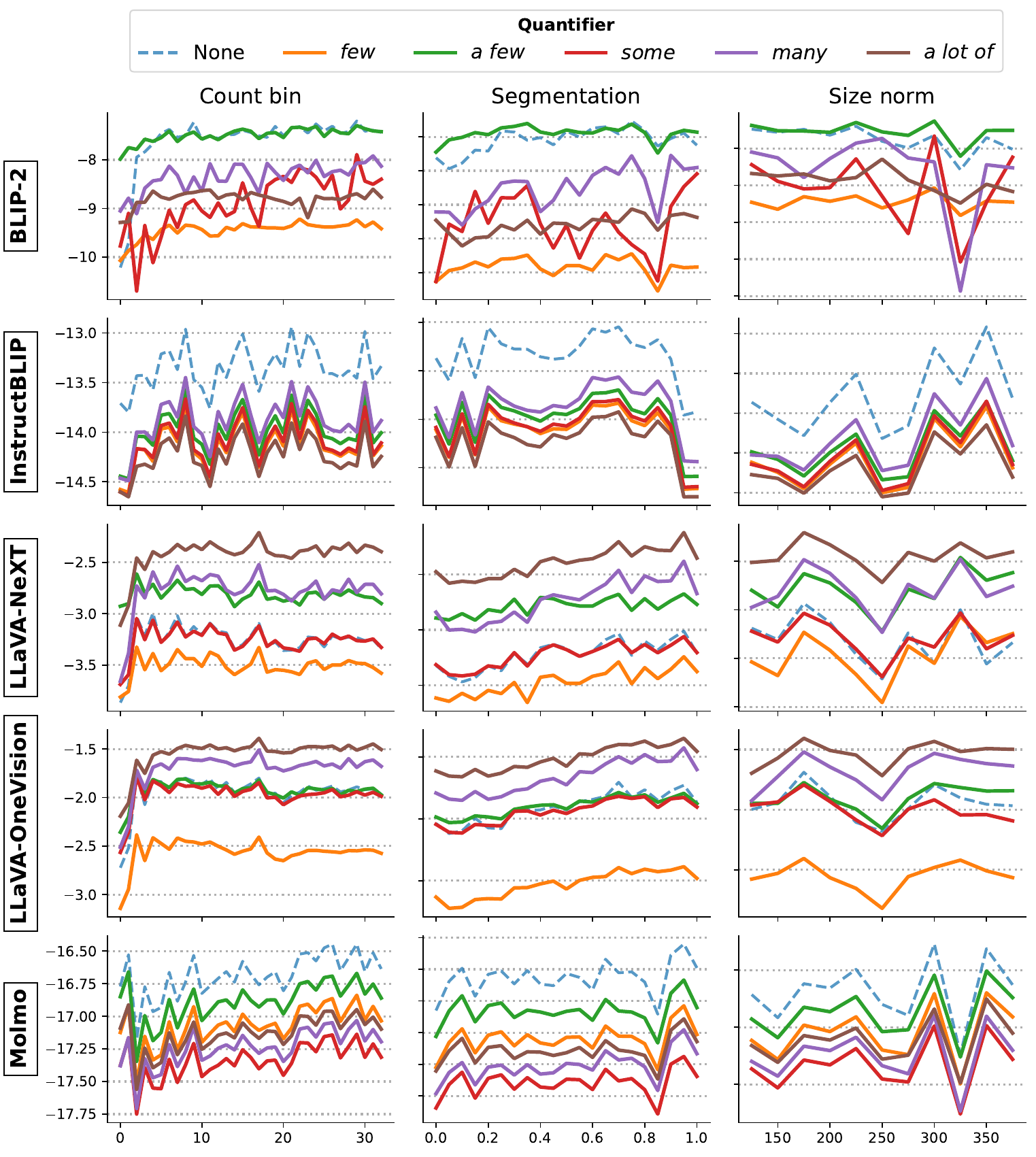}
    \caption{\textbf{Log probabilities extracted for statements as a response to ``How would you describe the image?''} The most obvious deviation from Figure~\ref{fig:exp1_logprobs} in \S\ref{sec:logprobs} are the plots for the two LLaVA models, that no longer appear to distinguish between the different quantified statements.}
    \label{fig:app_exp1_no_amounts}
\end{figure}
\section{Supplementary Material Experiment 1}\label{appendix:exp1}

\subsection{Targeting amounts}
In Figure~\ref{fig:app_exp1_no_amounts} we show the patterns of the VLMs across all predictors for the prompt that does \emph{not} target the amount. The question presented to the models is ``How would you describe the image?'', and we extract log probabilities for expressions of the form ``There are \verb|[QUANT]| \verb|[OBJECT]| in the image'' (unchanged from those used in \S\ref{sec:logprobs}).

For LLaVA-NeXT and LLaVA-OneVision, the two models observed in \S\ref{sec:logprobs} to have the highest correlation with human ratings, we now find that patterns are the same across all quantifiers. We now find a ``layered'' or ``stacked'' pattern that is indicative of a bias towards a specific quantifier: while LLaVA-NeXT and LLaVA-OneVision tend towards always responding with \textit{a lot of}, InstructBLIP and Molmo favor the unquantified statement.

\subsection{Spearman's Correlation}\label{appendix:exp1_spearman}
In \S\ref{sec:logprobs} Table~\ref{tab:exp1_correlations}, we computed Pearson's correlation because our analyses showed strong reasons to assume a linear relationship between human and model scores.
Moreover, we observed that in some settings, the data distribution is highly concentrated at one end of the scale (e.g. both human and model scores are clustered towards the higher scores for ``many'').
In such cases, rank-based correlations such as Spearman's can become unstable or less informative, since small variations in scores can change the overall ordering.

However, to provide a more complete analysis, we include Spearman’s scores in Table~\ref{tab:exp1_correlations_spearman}.
While Pearson’s captures linear agreement, Spearman’s provides additional insight into rank-order consistency between human and model judgments. Importantly, the overall trends and conclusions remain consistent across both metrics.

\begin{table}
    \centering
    \small
    \begin{tabular}{l|ccccc}
    \toprule
    \textbf{Model} & \textbf{\textit{few}} & \textbf{\textit{a few}} & \textbf{\textit{some}} & \textbf{\textit{many}} & \textbf{\textit{a lot of}} \\ \midrule
        BLIP-2          & \textbf{-0.18} & \textbf{-0.19} & \textbf{-0.07} & \textbf{0.25} & \textbf{0.13} \\
        InstBLIP    & 0.03 & 0.03 & -0.02 & 0.03 & -0.01 \\
        LLaVA-N      & \textbf{0.31} & \textbf{0.34} & \textbf{0.22} & \textbf{0.30} & \textbf{0.49} \\ 
        LLaVA-O & \textbf{0.29} & \textbf{0.39} & \textbf{0.21} & \textbf{0.36} & \textbf{0.42} \\
        Molmo           & \textbf{0.13} & \textbf{0.19} & \textbf{0.08} & \textbf{-0.19} & \textbf{-0.22} \\
    \bottomrule
    \end{tabular}
    \caption{\textbf{Spearman's correlation between human ratings and model log probabilities.} Numbers in boldface are statistically significant ($p <$ 0.05).}
 \label{tab:exp1_correlations_spearman}
 \end{table}

\begin{table*}[ht]
\centering
\small
\begin{tabular}{ll|c|c|ccccc}
\toprule
\multicolumn{2}{c|}{} 
    & \multirow{2}{*}{\textbf{Intercept}} 
    & \multirow{2}{*}{\textbf{Main}}
    & \multicolumn{5}{c}{\textbf{Quantifier}} \\
    &&&& \textbf{\textit{few}} & \textbf{\textit{a few}} & \textbf{\textit{some}} & \textbf{\textit{many}} & \textbf{\textit{a lot of}} \\
    \midrule

\multirow{4}{*}{\textbf{BLIP-2}}
    & \textbf{Main effect} & \multirow{4}{*}{\textbf{0.41}} & -- & \textbf{-0.89} & \textbf{-0.09} & \textbf{-0.79} & \textbf{-0.26} & \textbf{-1.37} \\
    & \textbf{Count}        && 0.03 & \textbf{0.21} & 0.02 & \textbf{-0.10} & -0.03 & 0.01 \\
    & \textbf{Segmentation} && 0.03 & \textbf{0.06} & -0.04 & -0.02 & \textbf{0.09} & -0.02 \\
    & \textbf{Size norm}    && 0.02 & 0.01 & -0.03 & \textbf{-0.07} & \textbf{-0.13} & 0.06 \\ 
    \midrule

\multirow{4}{*}{\textbf{InstructBLIP}}
    & \textbf{Main effect} & \multirow{4}{*}{\textbf{0.57}} & -- & \textbf{-0.76} & \textbf{-0.82} & \textbf{-0.86} & \textbf{-0.46} & \textbf{-1.20} \\
    & \textbf{Count}        && -0.02 & -0.02 & -0.01 & -0.01 & 0.03 & 0.00 \\
    & \textbf{Segmentation} && \textbf{-0.11} & -0.01 & 0.00 & 0.00 & 0.00 & 0.02 \\
    & \textbf{Size norm}    && \textbf{0.33} & \textbf{-0.08} & \textbf{-0.09} & \textbf{-0.06} & 0.02
        & \textbf{-0.09} \\
    \midrule

\multirow{4}{*}{\textbf{LLaVA-NeXT}}
    & \textbf{Main effect} & \multirow{4}{*}{\textbf{-0.86}} & -- & \textbf{-0.05} & \textbf{1.00} & \textbf{0.31} & \textbf{2.10} & \textbf{2.08} \\
    &\textbf{Count}         && \textbf{-0.12} & \textbf{-0.03} & \textbf{-0.07} & \textbf{-0.04} & \textbf{0.21} & \textbf{0.26} \\
    & \textbf{Segmentation} && \textbf{-0.12} & 0.00 & -0.03 & 0.02 & \textbf{0.14} & \textbf{0.13} \\
    & \textbf{Size norm}    && \textbf{-0.08} & \textbf{0.08} & \textbf{0.12} & \textbf{0.03} & \textbf{0.15} & \textbf{0.15} \\
    \midrule

\multirow{4}{*}{\textbf{LLaVA-OneVision}}
    & \textbf{Main effect} & \multirow{4}{*}{\textbf{-1.25}} & -- & \textbf{0.39} & \textbf{1.68} & \textbf{0.77} & \textbf{2.46} & \textbf{2.32} \\
    & \textbf{Count}        && \textbf{-0.09} & 0.00 & -0.01 & -0.02 & \textbf{0.22} & \textbf{0.22} \\
    & \textbf{Segmentation} && \textbf{-0.05} & -0.02 & -0.01 & 0.01 & \textbf{0.07} & \textbf{0.05} \\
    & \textbf{Size norm}    && \textbf{-0.05} & \textbf{0.04} & \textbf{0.05} & -0.03 & \textbf{0.12} & \textbf{-0.09} \\
    \midrule

\multirow{4}{*}{\textbf{Molmo}}
    & \textbf{Main effect} & \multirow{4}{*}{\textbf{0.73}} & -- & \textbf{-0.71} & \textbf{-0.97} & \textbf{-1.35} & \textbf{-0.85} & \textbf{-1.30} \\
    & \textbf{Count}        && \textbf{-0.11} & 0.03 & 0.03 & \textbf{-0.05} & 0.02 & 0.04 \\
    & \textbf{Segmentation} && \textbf{-0.19} & 0.00 & 0.00 & -0.01 & 0.02 & 0.03 \\
    & \textbf{Size norm}    && \textbf{0.22} & -0.01 & -0.04 & -0.05 & 0.01 & \textbf{-0.06} \\
    \bottomrule

\end{tabular}
\caption{\textbf{Linear Mixed Effects estimates for all VLMs tested.} We discuss the estimates for LLaVA-OneVision in \S\ref{sec:logprobs}.}
\label{tab:exp1_lmms_extra}
\end{table*}

\subsection{LMMs for all remaining models}
In Table~\ref{tab:exp1_lmms_extra}, we report estimates of LMMs for BLIP-2, InstructBLIP, LLaVA-NeXT and Molmo.

%% file: appendix/appendix_exp2_prompts.tex
\section{Supplementary Material Experiment 2}\label{appendix:exp2}

\subsection{Prompts for Score Generation}\label{appendix:exp2_prompts}
Below we list the 10 prompts that we have tested for Experiment~2. The prompt listed in boldface is discussed in \S\ref{sec:model_judgments}. This prompt is the one that the VLMs were least likely to interpret as a binary question. That is, we sought a prompt that explicitly encouraged the models to respond within a range of values, akin to the graded responses from human participants. Given that VLMs will generally have encountered multiple-choice question-answering prompts in training, we decided to mimic this format (following \citet{belem2024-uncertainty}) for our prompt in \S\ref{sec:model_judgments}.

\begin{enumerate}
    \item ``On a scale of 0 (completely inaccurate) to 100 (completely accurate), how accurate is the following statement for the image? Please only respond with a number between 0 and 100.\\
    \verb|[Statement]|''

    \item ``Question: How accurate is the following statement for the image? Respond only with a rating between 0 (completely inaccurate) and 100 (completely accurate).\\
    Statement: \verb|[Statement]| \\
    Answer: ''

    \item ``On a scale of 0 (completely inaccurate) to 100 (completely accurate), how accurate is the following statement for the image? Respond only with a number. Decimals are allowed. \\
    \verb|[Statement]|''

    \item ``How accurate is the statement for the image? Please only respond with a number between 0 and 100, where 0 is `completely inaccurate' and 100 `completely accurate'.\\
    \verb|[Statement]|''

    \item ``\textbf{On a scale of 0 (completely inaccurate) to 100 (completely accurate), how accurate is the following statement for the image? \\ Please respond with one of the following options: 0, 5, 10, 15, 20, 25, 30, 35, 40, 45, 50, 55, 60, 65, 70, 75, 80, 85, 90, 95, 100.}\\
    \verb|[Statement]|''

    \item ``How likely is the following caption given the image? Please respond with a number between 0 and 100, where\\- 0 is `not likely at all'\\- 100 is `highly likely'. \\
    Caption: \verb|[Statement]|''

    \item ``What is the probability that the following sentence matches the image?\\
    \verb|[Statement]|''

    \item ``What is the probability that the following sentence matches the image?\\
    Sentence: \verb|[Statement]| \\
    Answer: ''

    \item ``What is the probability that the following sentence matches the image? Please only respond with a number between 0 and 100. \\
    \verb|[Statement]|''

    \item ``What is the probability that the following sentence matches the image? Please only respond with a number between 0 and 1. \\
    Sentence: \verb|[Statement]| \\
    Answer: ''

\end{enumerate}

\subsection{Distribution of Generated Scores}
\begin{figure}[t]
    \centering
    \includegraphics[width=0.58\linewidth]{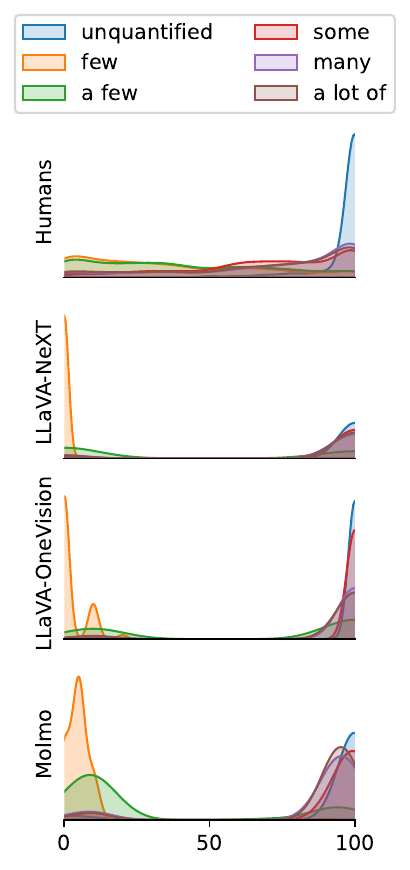}
    \caption{\textbf{Distributions for human ratings and scores generated by VLMs per quantifier.}}
    \label{fig:app_exp2_distribution}
\end{figure}

Figure~\ref{fig:app_exp2_distribution} shows density plots displaying the distributions of human ratings in \dataset, as well as scores generated by VLMs as a response to prompt 5 in Appendix~\ref{appendix:exp2_prompts}, discussed in \S\ref{sec:model_judgments}. Note that for LLaVA-NeXT, LLaVA-OneVision and Molmo, the scores tend towards the extremes. However, in the human distribution, this is only the case for the unquantified control statement (as expected).

%% file: appendix/appendix_exp3.tex
\section{Supplementary Material for Experiment 3}\label{appendix:exp3}
Similar to what we did in Appendix~\ref{appendix:exp1_spearman} for Experiment~1, we now provide Spearman's correlation in Table~\ref{tab:exp3_correlations_spearman} corresponding to the relations in Table~\ref{tab:exp3_correlations} for Experiment~3 (\S\ref{sec:exp3}).
We discuss the rationale for doing so in Appendix~\ref{appendix:exp1_spearman}.

\begin{table}
\centering
\small
\setlength{\tabcolsep}{5pt}

\begin{tabular}{ll|ccccc}
\toprule
&& \textbf{\textit{few}} & \textbf{\textit{a few}} & \textbf{\textit{some}} & \textbf{\textit{many}} & \textbf{\textit{a lot of}}\\\midrule

\multirow{2}{*}{\rotatebox[origin=c]{90}{\textbf{INB}}}
    & $r(\textup{VAQ})$ & -0.01 & -0.02 & 0.01 & -0.01 & 0.05 \\
    & $r(\textup{EXP1})$ & \textbf{-0.11} & \textbf{-0.12} & \textbf{-0.10} & \textbf{-0.11} & \textbf{-0.13}\\
    \midrule

\multirow{2}{*}{\rotatebox[origin=c]{90}{\textbf{LLN}}} 
    & $r(\textup{VAQ})$ & \textbf{0.30} & \textbf{0.28} & \textbf{0.15} & \textbf{0.26} & \textbf{0.28}\\
    & $r(\textup{EXP1})$ & \textbf{0.37} & \textbf{0.36} & \textbf{0.27} & \textbf{0.29} & \textbf{0.25}\\
    \midrule
\multirow{2}{*}{\rotatebox[origin=c]{90}{\textbf{LLO}}} 
    & $r(\textup{VAQ})$ & \textbf{0.39} & \textbf{0.39} & \textbf{0.19} & \textbf{0.28} & \textbf{0.36}\\
    & $r(\textup{EXP1})$ & \textbf{0.32} & \textbf{0.42} & \textbf{0.26} & \textbf{0.15} & \textbf{0.25} \\
    \midrule
\multirow{2}{*}{\rotatebox[origin=c]{90}{\textbf{MOL}}} 
    & $r(\textup{VAQ})$ & \textbf{0.21} & \textbf{0.28} & \textbf{0.17} & \textbf{0.20} & \textbf{0.26} \\
    & $r(\textup{EXP1})$ & \textbf{0.23} & \textbf{0.27} & \textbf{0.23} & \textbf{-0.06} & \textbf{-0.11}
    \vspace{1.25pt}\\
\bottomrule

\end{tabular}
\caption{\textbf{Spearman's $\bm{\rho}$ of log probabilities in Experiment~3 with human data (VAQ) and log probabilities from Experiment~1 (EXP1)}. Models shown are InstructBLIP (\textbf{INB}), LLaVA-NeXT (\textbf{LLN}), LLaVA-OneVision (\textbf{LLO}) and Molmo (\textbf{MOL}). Boldfaced numbers are statistically significant.}
\label{tab:exp3_correlations_spearman}
\end{table}

%% file: appendix/appendix_licenses.tex
\section{Dataset Licenses}
For the construction of the \dataset dataset, we have used images from several existing datasets. These datasets are released under open-source or permissive licenses:
Apache License 2.0 (TallyQA), MIT License (FSC-147/FSC-133), and Creative Commons Attribution 4.0 International License (Visual Genome and VQA/VQA2).
Our use of these resources is consistent with the terms and intended scope of their respective licenses.